\documentclass[10pt,twocolumn,letterpaper]{article}

\usepackage{cvpr}
\usepackage{times}
\usepackage{epsfig}
\usepackage{graphicx}
\usepackage{amsmath}
\usepackage{amssymb}
\usepackage{float}
\usepackage{tabularx}
\usepackage{caption}
\usepackage[export]{adjustbox}
\usepackage[super]{nth}
\usepackage[pagebackref=true,breaklinks=true,letterpaper=true,colorlinks,bookmarks=false]{hyperref}

\cvprfinalcopy 

\def\cvprPaperID{****} 

\ifcvprfinal\fi
\begin{document}

\title{GANiry: Bald-to-Hairy Translation Using CycleGAN}

\author{Fidan Samet \qquad Oguz Bakir \\
Department of Computer Engineering, Hacettepe University\\
{\tt\small \{fidanlsamet,oguzbakir0\}@gmai1.com}
}


\maketitle

\begin{abstract}
   This work presents our computer vision course project called bald men-to-hairy men translation using CycleGAN. On top of CycleGAN architecture, we utilize perceptual loss in order to achieve more realistic results. We also integrate conditional constrains to obtain different stylized and colored hairs on bald men. We conducted extensive experiments and present qualitative results in this paper. Our code and models are available at \url{https://github.com/fidansamet/GANiry}.
\end{abstract}


\section{Introduction}

\subsection{Problem \& Aim}

Whether the bald men like to be bald or not, sometimes they wonder how they would look if they had hair. To that end, bald-to-hairy translation can be performed quickly and realistically by using deep learning methods. We focus on image-to-image translation problem for the bald-to-hairy translation task to add hair to bald men in the context of this project.

We aim to create mapping between bald men and hairy men images to generate hair to bald men. As a result of our research, we found out that the most effective methods for this task are Generative Adversarial Networks (GAN) based methods. By considering the literature, we can categorize GAN into two methods which are paired and unpaired methods. For our problem, there are two datasets. One dataset is for bald men images which is source dataset and the other one is for hairy men images which is target dataset. Therefore, the unpaired approach must be considered for bald-to-hairy translation problem.

After obtaining the baseline results, we try to improve the baseline model by adding conditional layer and perceptual loss so that it can generate four different colors of hair which are black, brown, blond, gray and two different styles of hair which are straight, wavy.


\subsection{Challenges}
\label{sec:challenges}

We can collect the challenges of this project into three groups: dataset, classification and time.

For dataset, there are three main challenges:

\begin{itemize}
	\item There are two datasets in our project, one for bald men and the other one for hairy men. Since there is no dataset specifically prepared for our problem, we prepared both datasets ourselves from the existing ones.
	\item Perhaps the biggest challenge of this project is that it relies on unpaired data. We cannot consider this task on paired data, because there does not exist any paired dataset prepared for our translation problem in the literature. So we need to work on unpaired data and working with unpaired data is a challenge.
	\item Our hairy dataset contains large number of images. However, our bald dataset is not as large as the hairy dataset. We cannot train all of the images in datasets due to resource and time issues. Therefore, we randomly sample them, but this is an obstacle to our best performance.
\end{itemize}

Hair classification is another challenge. In the second phase of this project, we classify and generate the hairs conditionally. The classes that has poor image numbers on hairy dataset create challenge for classifying and generating hairs of that class. This challenge can change accuracy of obtained results. Also, classifying unpaired data is an another challenge. Therefore, the generated hair results depend on this issue.

Last challenge is the time. Training GAN based methods takes long time. It takes about one day to train a model we have determined and get the result. If the model we train does not give good results, the time consequences are heavy. This becomes even bigger challenge as we try to tune the hyperparameters to obtain the best model.


\subsection{Method}

Since our project is based on unpaired data, we use CycleGAN~\cite{zhu2017unpaired} as our baseline. According to our research, it is one of the most advanced GAN methods and it can perform image-to-image translation on unpaired data. So our method rely on CycleGAN.

We add conditional layer to our baseline for classifying and generating four different colors of hair which are black, brown, blond, gray and two different styles of hair which are straight, wavy conditionally. Since we do not have enough computing power and time, we keep the number of training images low. Therefore, at first we experiment with four hair classes (black and blond hair colors, straight and wavy hair styles) which can be distinguished better and get more distinctly detachable results. Also, we experiment different hyperparameters such as number of training images and image sizes to get better results. Then we try to generate all six hair classes and experiment it by image size hyperparameter. We try different baselines and obtain the the model that gives the best results on our task.

After obtaining the best baseline by deciding the hyperparameters such as class number, image number and image size, we try to make improvements on results for bald-to-hairy translation task. Firstly, we add perceptual loss beside cycle-consistency loss and GAN losses of CycleGAN model to improve the results. Then we experiment removing cycle-consistency loss and adding perceptual loss.

After deciding perceptual loss addition and cycle-consistency loss removal, we try a different generator architecture, U-Nets, to obtain better results. At first, we experiment it without adding conditional layer to compare with ResNet generator architecture without conditional layer. Then we experiment adding conditional layer to the generator architecture.

Finally, after obtaining the best generator architecture, we experiment the best model with the maximum number of training images to get the best results.


\subsection{Related Works}

According to our research, there are several works related to ours. The most important of these works is CycleGAN. It performs unpaired image-to-image translation by using cycle consistent adversarial networks and performs general generation. It can perform translation between horses and zebras, summer and winter, painting and photos, apples and oranges, etc. It is an important work for our task since we use it as baseline.

StyleGAN\cite{karras2019style} can do any styling with given dataset. That includes human hair styles, too. However, there is no information about bald hairstyles or removing styles.

AttGAN\cite{he2019attgan} can do translations between different facial attributes such as hair styles, beard styles, mouth positions, etc. Nevertheless, it has really low accuracy(~35\%) on bald images and can do big failures in transformations.

Hair-GANs\cite{zhang2019hair} can detect 3D shape of hair from single 2D image, but there is no information about adding nor removing hair from images. It can only compute the 3D volumetric field as the structure guidance for the final hair synthesis.

The fundamental difference of our project from these related works is that we add hair specifically to the bald men. So we mainly focus on bald-to-hairy translation task.


\section{Method}

\subsection{Definitions}
\textbf{Deep Learning:} It is the field involving artificial neural networks containing layers and hidden layers, similar to machine learning algorithms in artificial intelligence. Deep learning operations require lots of learning, data and time. It has supervised, semi-supervised and unsupervised categories.

\textbf{Unpaired Data:} Data is required for deep learning training. It is described as unpaired or independent when the sets of data arise from separate individuals.

\textbf{GAN(Generative Adversarial Networks):} It is a subclass of machine learning frameworks. Two neural networks compete with each other in a game. It is used to generate new data with the same statistics as the training set. We will be using GAN for image-to-image translation using bald and hairy men images.

\textbf{Image-to-Image Translations:} It is a subclass of computer vision that can learn the mapping between two images. In our project, we will be doing image-to-image translation between bald and hairy men images.

\textbf{CycleGAN:} It is one of the most advanced GAN methods. It performs unpaired image-to-image translation by using cycle-consistent adversarial networks. We use CycleGAN as our baseline.

\textbf{Perceptual Loss:} It sums all squared errors between all pixels and takes the mean of it. It is often used to compare two different images that look almost the same. It shows errors in generated images.


\subsection{Approach}

The first step of our method for solving bald-to-hairy translation problem is creating a model based on CycleGAN. Next, we try different baseline models. We decide the best baseline according to the results we have received. After choosing the baseline, we consider four things for improvements:

\begin{figure*}[t!]
\centering
\begin{minipage}{.5\textwidth}
  \centering
  \includegraphics[width=1.0\linewidth]{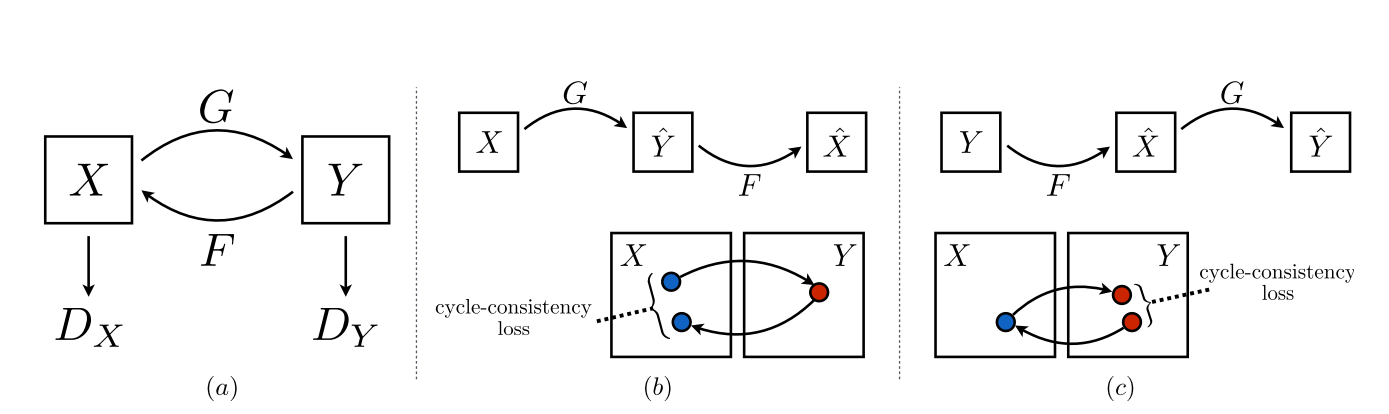}
  \captionof{figure}{Model of CycleGAN\cite{zhu2017unpaired}}
  \label{fig:cyclegan_model}
\end{minipage}%
\begin{minipage}{.5\textwidth}
  \centering
  \includegraphics[width=1.0\linewidth]{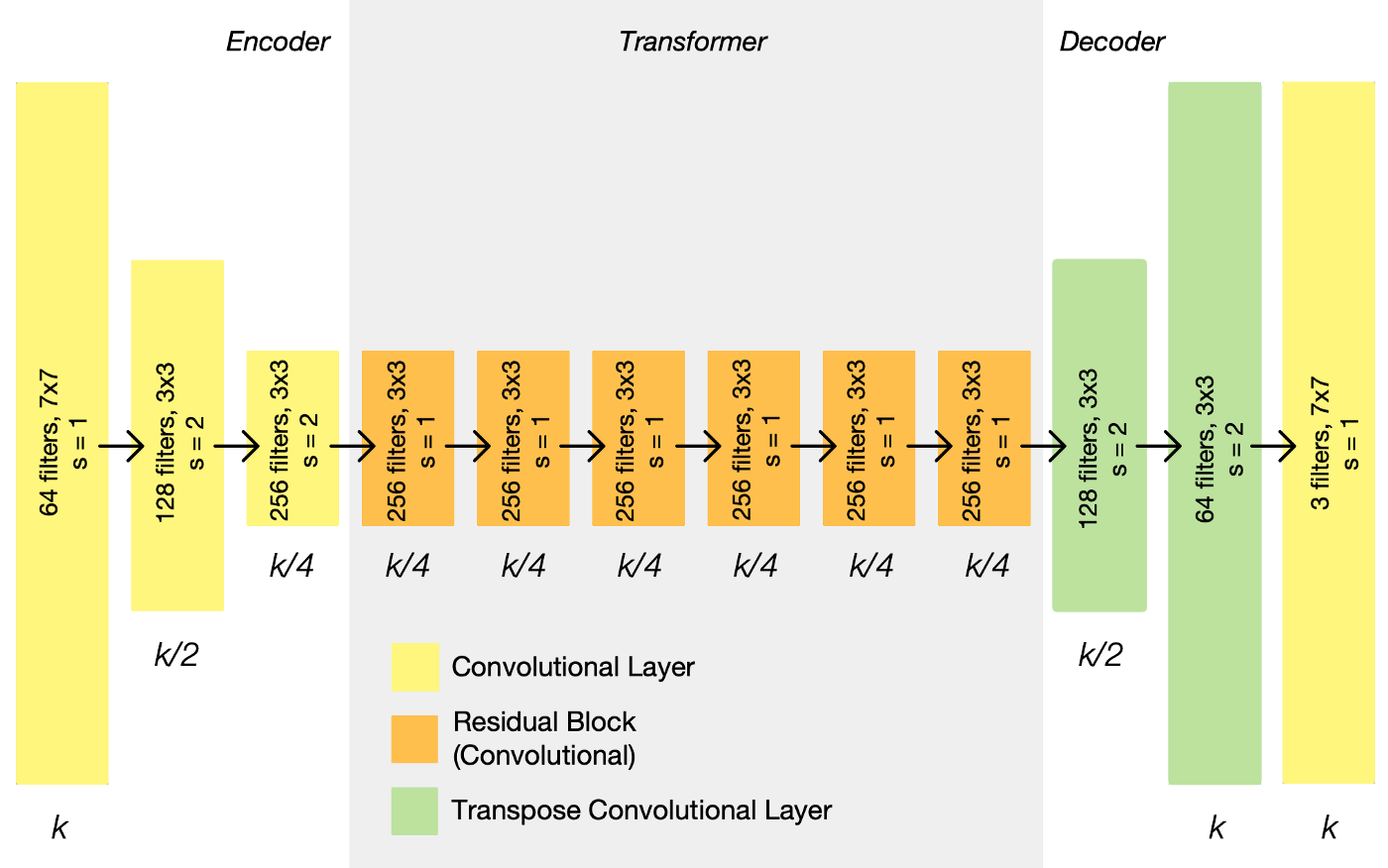}
  \captionof{figure}{Architecture of CycleGAN\cite{zhu2017unpaired}}
  \label{fig:cyclegan_arch}
\end{minipage}
\end{figure*}

\begin{itemize}
	\item Adding conditions to CycleGAN for generating different hair colors and styles.
	\item Adding perceptual loss~\cite{zhang2018unreasonable} in addition to cycle consistency and GAN losses to improve the results since there is no perceptual loss in the current version of CycleGAN.
	\item Trying U-Net generator architecture instead of ResNet generator architecture to observe different generator architecture impacts on our models.
	\item Selecting the best generator architecture and do final testing with the best hyperparameters.
\end{itemize}

As Figure \ref{fig:cyclegan_model} shows, there are two domains X and Y, two generators G and F, two discriminators Dx and Dy in our model as CycleGAN model. In the sub-figure a, cycles are shown. In the sub-figure b and sub-figure c, the detailed views of cycles are given. The main idea of this model is when G generator maps X domain to Y domain, F generator maps the resulted Y domain back to X domain by using the discriminators. The cycle-consistency loss is calculated according to X domain and X domain obtained by F generator.

By calculating the cycle-consistency loss in training, model becomes more cycle-consistent. So X domain and X domain obtained by F generator begins to be almost the same. The sub-figure b shows X to Y translation and sub-figure c shows Y to X translation.

Since we use this cycle model, we also obtain hairy-to-bald translation results. However, our aim is not to improve hairy-to-bald translation, but to improve bald-to-hairy translation results.

Figure \ref{fig:cyclegan_arch} shows architecture of CycleGAN. This network contains two stride-2 convolutions, several residual blocks and two fractionally strided convolutions with stride 1/2. There are 6 blocks for 128$x$128 images and 9 blocks for 256$x$256 and higher resolution training images. Instance normalization is used. For the discriminator networks there are 70$x$70 PatchGANs which aim to classify whether 70$x$70 overlapping image patches are real or fake. Such a patch-level discriminator architecture has fewer parameters than a full-image discriminator and can work on arbitrarily-sized images in a fully convolutional fashion.~\cite{zhu2017unpaired}

By adding condition layer to CycleGAN, we can generate hair with colors and styles. We create the condition layer using an embedding layer and combining it with the data in generator and discriminator blocks. Then it is fed into CycleGAN model.

For further improvements on generations, we can add perceptual loss and experiment with and without cycle-consistency loss. As perceptual loss~\cite{zhang2018unreasonable} is proposed, after passing input and cycle output from VGGNet16, we can obtain perceptual loss by giving them to the loss function. According to the obtained results, we can use the best option for next experiments.

If still there is no strict distinction between different hair colors and styles, we can change ResNet generator architecture with U-Net generator architecture without any condition layer. The goal of that change is to see differences between ResNet and U-Net generator architectures impact on our generations. If U-Net generator architecture results are better than ResNet generator architecture, we select U-Net as the our generator and do a final training using all 4430 train images.

\section{Experimental Settings}

\subsection{Dataset}

Since we use GAN to generate hairy men images, we need a lot of annotated men images to train the model. CelebA\cite{liu2018large} dataset fits for our task. Dataset information can be found in Table \ref{table:dataset}. We use eye-aligned version of this dataset for hairy and bald men data. Due to hardware and time issues, we subsample 1000, 2000 and 4430 training images
and separate 100 test images for each domain.

To create hair classes, we filter images with necessary annotations (black, blond, brown, gray, straight and wavy hair). Table \ref{table:hair_class} shows the distribution over these classes.

\begin{table}
\begin{center}
\begin{tabular}{|l|l|l|l|}
\hline
\textbf{\# images} & \textbf{Total} & \textbf{Hairy Men} & \textbf{Bald Men} \\
\hline\hline
CelebA & 202,599 & 79904 & 4530 \\
\hline
\end{tabular}
\caption{Number of Images Distribution Over Dataset}
\label{table:dataset}
\end{center}
\end{table}

\begin{table}
\begin{center}
\begin{tabular}{|l|l|}
\hline
\textbf{Hair Class} & \textbf{\# of images} \\
\hline\hline
Black & 25156 \\
Blond & 1749 \\
Brown & 12788 \\
Gray & 7235 \\
Straight & 20471 \\
Wavy & 11892 \\
\hline
\end{tabular}
\caption{Hair class distribution over dataset}
\label{table:hair_class}
\end{center}
\end{table}


\subsection{Experimental Setup}

Our physical system has NVIDIA GTX 1060 with 6GB vRAM and 1280 CUDA cores. After some time, we switched to dual NVIDIA GTX1080TI with 22GB vRAM and 7168 CUDA cores in total.


Our training data consists of 256$x$256 pixel images. Since this is an issue of performance for us, we downscale the image size to 64$x$64 and 128$x$128.


\subsection{Implementation}

\begin{table*}[ht]
\begin{tabularx}{\textwidth}{| X | X | X | X | X | X | X | X |}
\hline
Experiment Number & Generator Arch & \# Images & Image Size & \# Classes & Loss & Train Time & GPU \\
\hline\hline
1 & ResNet-9 & 1000 & 256$x$256 & - & C & 36 h 45 m & GTX1060 \\ 
2 & ResNet-6 & 1000 & 64$x$64 & 4 & C & 8 h 35 m & GTX1060 \\
3 & ResNet-6 & 4430 & 64$x$64 & 4 & C & 35 h 35 m & GTX1060 \\
4 & ResNet-6 & 2000 & 64$x$64 & 4 & C & 15 h 45 m & GTX1060 \\
5 & ResNet-6 & 2000 & 128$x$128 & 4 & C & 24 h 41 m & GTX1060 \\
6 & ResNet-6 & 2000 & 64$x$64 & 6 & C & 15 h 22 m & GTX1060 \\
7 & ResNet-6 & 2000 & 128$x$128 & 6 & C & 26 h 36 m & GTX1060 \\

8 & ResNet-6 & 2000 & 128$x$128 & 4 & C + P & 14 h 49 m & GTX1080TI \\
9 & ResNet-6 & 2000 & 128$x$128 & 4 & P & 16 h 52 m & GTX1080TI \\
10 & U-Net-128 & 2000 & 128$x$128 & - & C & 20 h 17 m & GTX1060 \\
11 & U-Net-128 & 2000 & 128$x$128 & 4 & P & 14 h 57 m & GTX1080TI \\
12 & ResNet-6 & 4430 & 128$x$128 & 4 & P & 29 h 51 m & GTX1080TI \\

\hline
\end{tabularx}
\caption{All experiments with 200 epoch (C: Cycle-consistency Loss, P: Perceptual Loss)}
\end{table*}

Every experiment is done in light of previous experiment results. If the changes make improvements on results, this incident is used on next experiments. The first experiment has vanilla CycleGAN hyperparameters and they are changed after the experiment results.

\textbf{\nth{1} Experiment:} In our first experiment, we used default hyperparameters of CycleGAN such as image size, generator architecture and losses. There was no condition layer for hair classes and with these parameters we cannot generate hairs with the given hair colors and styles.

\textbf{\nth{2} Experiment:} In this experiment, we decreased image size to 64$x$64 for faster training. By training faster, we can observe hyperparameter changes without losing too much time and we added condition layer with 4 classes which are black, blond, straight and wavy hair. But we could not achieve good condition separation with 1000 images.

\textbf{\nth{3} Experiment:} After bad results with 1000 images, we increased number of images to 4430 which is all bald images that we have. There was some good separation between conditions but the train time was too long. This experiment was the longest one.

\textbf{\nth{4} Experiment:} Because of the long training time with 4430 images we decreased number of images to 2000 and kept other hyperparameters as they are. The results did not change too much due to low image count. But all results were blurry due to low image resolution.

\textbf{\nth{5} Experiment:} To eliminate blurry images, we increased image size to 128$x$128. The train time increased by \%80 but obtained results have better clarity now.

\textbf{\nth{6} Experiment:} After finding the best hyperparameters, we tried to classify and generate more hair colors so we increased the number of classes to 6(Black Hair, Blond Hair, Brown Hair, Gray Hair, Straight Hair and Wavy Hair). The obtained results were blurry with 64$x$64 images and there was a bad classification between 6 classes.

\textbf{\nth{7} Experiment:} We increased image size to 128$x$128 for clear images. Since there are not enough images to classify 6 classes, the network could not classify 6 classes. Because of that we continued with 4 classes for condition layer.

With the experiments we have done up to this point, we chose all the best hyperparameters and the best model for our hair generation task. We added perceptual loss for better hair generation.

\textbf{\nth{8} Experiment:} In this experiment, we added perceptual loss in addition to cycle-consistency and GAN losses. The obtained results were better than without perceptual loss.

\textbf{\nth{9} Experiment:} After adding perceptual loss to our model, we removed cycle-consistency loss and kept perceptual loss in addition to GAN losses. The obtained results were better than the results of \nth{8} experiment, so we kept perceptual loss and removed cycle-consistency loss.

\textbf{\nth{10} Experiment:} After adding perceptual loss for better hair generation, classification and generation of different hair classes were not impressive because of that we changed ResNet with U-Net generator architecture. But for initial testing we experimented with initial CycleGAN losses and we did not add any condition layer. The obtained results were similar to ResNet results and U-Net generator models allocate 5 times more disk space in this experiment.

\textbf{\nth{11} Experiment:} To test U-Net generator architecture performance with condition layer we added 4 classes to our model and the obtained results were not impressive. Due to insufficient disk space usage and inconsistent results compared to ResNet generator architecture, we did not select U-Net as the best generator architecture.

\textbf{\nth{12} Experiment:} After obtaining the best hyperparameters and the best generator architecture, we experimented with all train images for best results of our model.

\section{Discussions and Conclusions}

Example inputs and outputs from our experiments could be found at the end of the paper.

\subsection{CycleGAN vs Conditional CycleGAN}

In our project, at first we used CycleGAN model as our baseline and obtained the first results. CycleGAN generated hair to bald men but it was inconsistent. The model generated hair in some images while did not generate in others. Also the generated hairs were not realistic, they seemed as if painted with brush.

To improve the initial baseline so that it can generate different hair colors and different hair styles, we added a condition layer. After experimenting different hyperparameters such as number of train images and image sizes with conditional CycleGAN, we tuned hyperparameters and obtained more consistent results. Although we obtained more consistent results, the conditions did not work very well. Since there was not enough data to provide each of the conditions, our conditioning method was not very effective. In most experiments we sampled 2000 images for training. The dominant color was black in that sampled dataset, so while black hair color learned well, there were not enough data to learn other hair colors. Also the sampled dataset was not enough to learn the exact map of different hair styles. So while our aim was to classify and generate 6 hair classes, we can classify and generate at most 4 hair classes(Black Hair, Blond Hair, Straight Hair, Wavy Hair) in a distinct way.

\subsection{Cycle-consistency vs Perceptual Loss}
CycleGAN uses cycle-consistency loss for cycle consistency. The generated hairs with cycle-consistency loss did not seem real, it seemed like they had been put with photoshop. By adding perceptual loss in addition to cycle-consistency loss and GAN losses, we obtained more clear and real generated hairs. We then experimented only perceptual loss in addition to GAN losses by removing cycle-consistency loss, the obtained results were better so that it was difficult to distinguish whether it was real hair or not.

\subsection{ResNet vs U-Net Generator Architecture}
U-Net is generally used in Pix2Pix. So it gives better results in paired data. When we experimented with U-Net generator architecture, we obtained a bad cycle consistency in results. It was because of the unpaired data. U-Net without conditional layer results were good but there was no conditional layer. Because of that we could not generate images we would like. After adding conditional layer to U-Net model we observed that it could not create distinctions between conditions and the generated images were sometimes better than ResNet results but usually they were not as good as ResNet results. Finally saved U-Net generator architecture models were allocating too much disk space due to depth of its network. So we decided to use ResNet generator architecture. With ResNet generator architecture, we obtained better and consistent results. The residual connections help improving the hair generation.

\subsection{Conclusions}
In general, our method for hair generation is succeeded since it can use all train images effectively to learn for generating hair. But our conditioning method is partially failed due to inefficiency of conditionally generated hairs.

CycleGAN was a good choice as a baseline for our task but there were some edge case problems in the results. For example when we trained light-skinned men and gave a dark-skinned man as a test image, we observed that the skin color in the test image became lighter. Also when we tested some experiments with old man images, our model made those men younger.

\begin{figure}[H]
\begin{center}
\fbox{
   \includegraphics[width=0.8\linewidth]{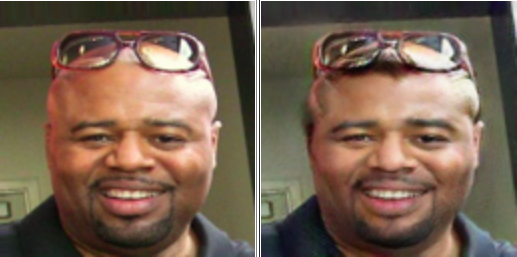}
}
\end{center}
   \caption{The result of bald man input with object on his head from \nth{9} experiment}
\label{fig:example}
\end{figure}

Finally, we can specify some limitations based on the results we have obtained. On our tests we found out that if there is any object on bald men heads such as glasses or hats, our models cannot generate hairs in these images. An example to this issue can be seen in Figure \ref{fig:example}. To eliminate this issue, more hairy and bald men images with an object on their heads needed on train dataset.

Also, if the bald dataset sampled for training does not contain enough number of completely bald men, the model does not learn well enough and when a completely bald man is given as a test image, the model cannot generate realistic hair.


\section{Improvements}

To improve the results, we can first consider increasing the number of train images. Our bald dataset is composed of 4530 images. We sample 100 images for test dataset. Therefore we have 4430 image for train dataset. Since CycleGAN needs the same number of images for source and target datasets, the maximum image number we can experiment with is 4430. By increasing the number of bald images, we can increase data for training. Hereby, the number of images for each condition increases, the model learns conditions better and the results improve.

As we discussed in \hyperref[sec:challenges]{\textit{Challenges section}}, the biggest challenge of this project is that it is based on unpaired data. Since there is not any paired dataset (bald and hairy images for the same men) prepared for this task, we work on unpaired data. This is an obstacle for our best performance goals. We could get more successful results with paired data.

Maybe we could experiment another state-of-the-art methods beside CycleGAN. We could experiment with baselining other methods which are more suitable for our task.

The best condition method can be found by trying different condition layers. A method more effective than the current one can be applied to obtain better results.

There are some issues on our dataset. The images in dataset are not very well annotated. There are woman images annotated as man, straight hair images annotated as wavy hair, etc. Also the images annotated as wavy and straight are not very reliable for men images, but reliable for woman images. Using dataset consists of more reliable annotated images could improve the results.

Finally, increasing the image size provides the model to learn better and the results to be improved.


\begin{figure*}
\captionsetup[subfigure]{labelformat=empty}
\centering
\setlength\tabcolsep{0pt} 
\hspace*{-1.0cm}%
\begin{tabular}{cccc}

\includegraphics[width=0.29\textwidth,  ,valign=m, keepaspectratio,]{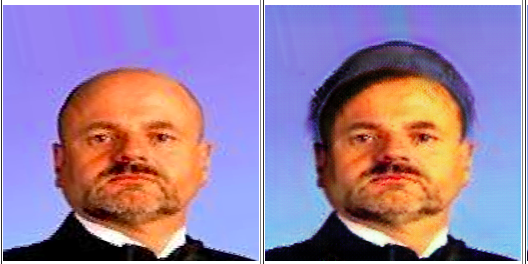} & 
\includegraphics[width=0.29\textwidth,  ,valign=m, keepaspectratio,]{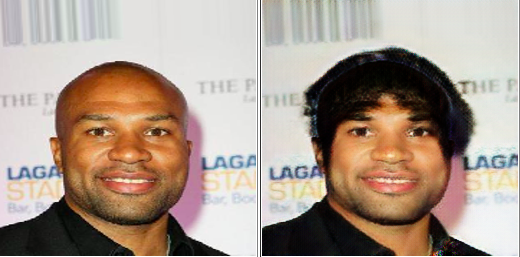}&
\includegraphics[width=0.29\textwidth,  ,valign=m, keepaspectratio,]{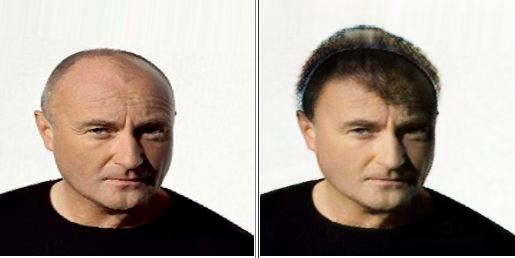}&
\includegraphics[width=0.29\textwidth,  ,valign=m, keepaspectratio,]{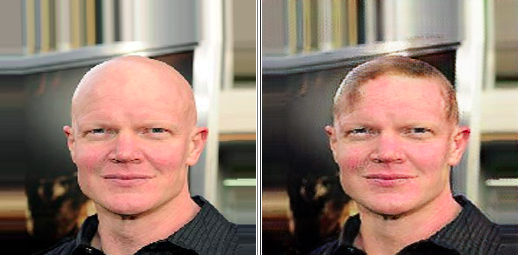}\\

\includegraphics[width=0.29\textwidth,  ,valign=m, keepaspectratio,]{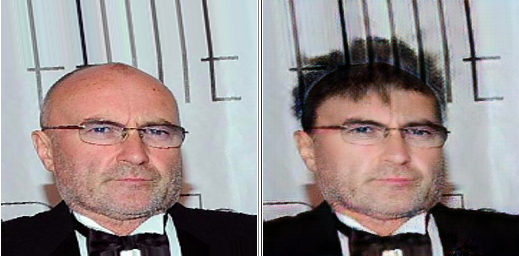} & 
\includegraphics[width=0.29\textwidth,  ,valign=m, keepaspectratio,]{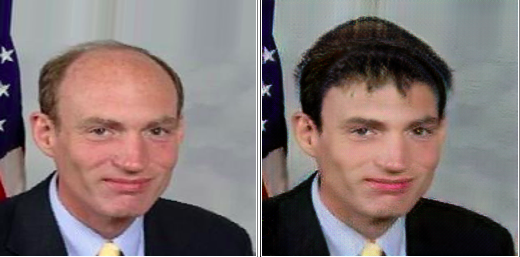}&
\includegraphics[width=0.29\textwidth,  ,valign=m, keepaspectratio,]{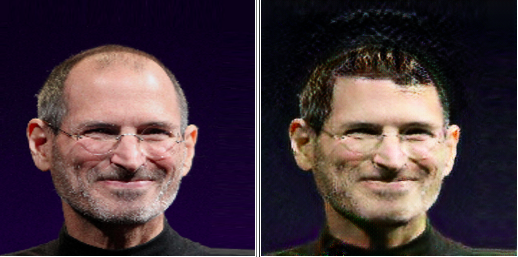}&
\includegraphics[width=0.29\textwidth,  ,valign=m, keepaspectratio,]{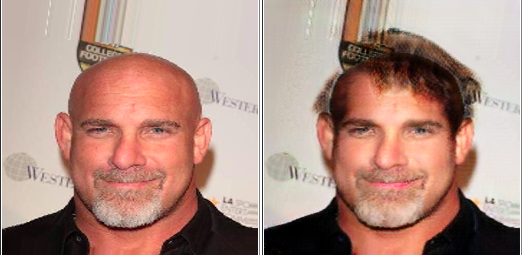}\\

\end{tabular}
\caption{Example results of \nth{1} experiment (Note that there is no condition layer in this experiment)}
\end{figure*}

\begin{figure*}
\captionsetup[subfigure]{labelformat=empty}
\centering
\setlength\tabcolsep{0pt} 
\hspace*{-1.0cm}%
\begin{tabular}{cccc}
\hspace*{0.5cm}Black-Straight & \hspace*{0.4cm}Black-Wavy & \hspace*{0.4cm}Blond-Straight & \hspace*{0.3cm}Blond-Wavy\\

\includegraphics[width=0.29\textwidth,  ,valign=m, keepaspectratio,]{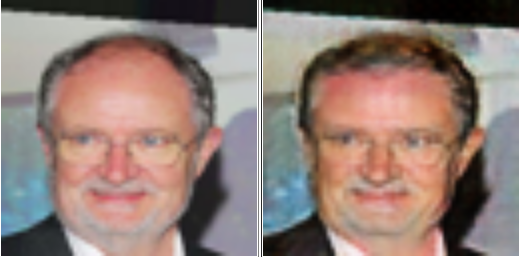} & 
\includegraphics[width=0.29\textwidth,  ,valign=m, keepaspectratio,]{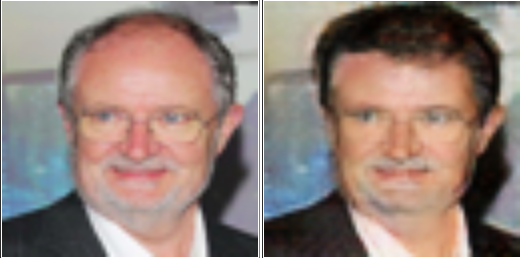}&
\includegraphics[width=0.29\textwidth,  ,valign=m, keepaspectratio,]{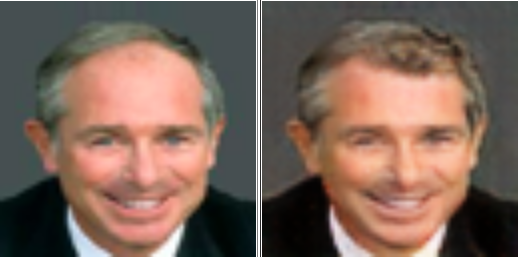}&
\includegraphics[width=0.29\textwidth,  ,valign=m, keepaspectratio,]{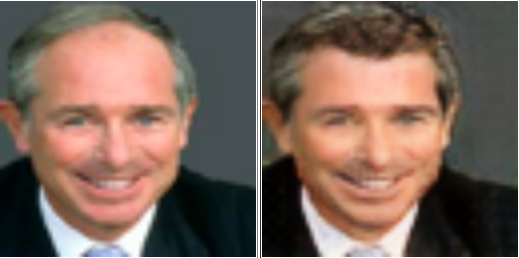}\\

\includegraphics[width=0.29\textwidth,  ,valign=m, keepaspectratio,]{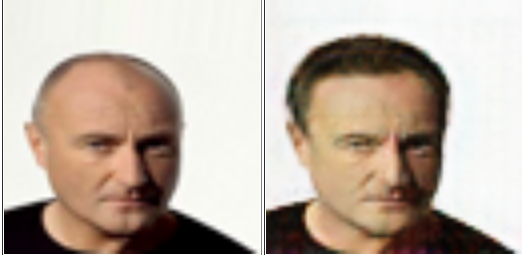} & 
\includegraphics[width=0.29\textwidth,  ,valign=m, keepaspectratio,]{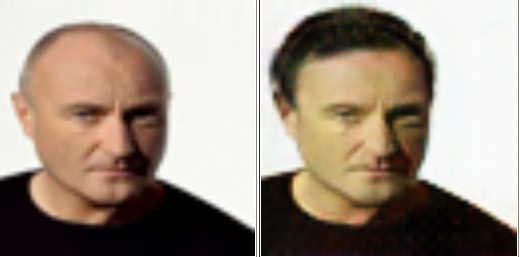}&
\includegraphics[width=0.29\textwidth,  ,valign=m, keepaspectratio,]{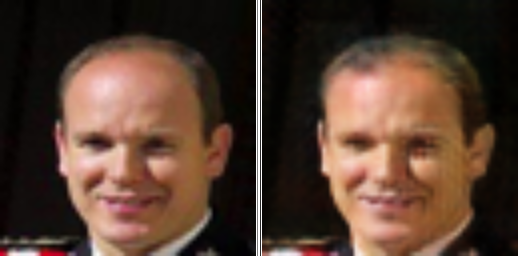}&
\includegraphics[width=0.29\textwidth,  ,valign=m, keepaspectratio,]{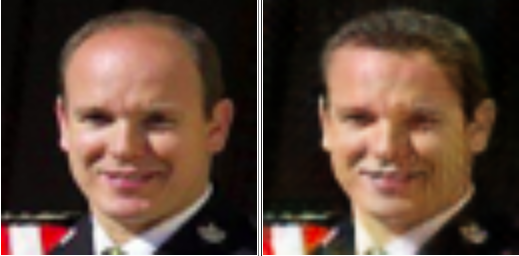}\\

\end{tabular}
\caption{Example results of \nth{2} experiment}
\end{figure*}

\begin{figure*}
\captionsetup[subfigure]{labelformat=empty}
\centering
\setlength\tabcolsep{0pt} 
\hspace*{-1.0cm}%
\begin{tabular}{cccc}
\hspace*{0.5cm}Black-Straight & \hspace*{0.4cm}Black-Wavy & \hspace*{0.4cm}Blond-Straight & \hspace*{0.3cm}Blond-Wavy\\

\includegraphics[width=0.29\textwidth,  ,valign=m, keepaspectratio,]{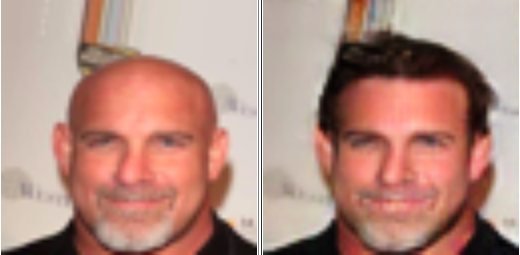} & 
\includegraphics[width=0.29\textwidth,  ,valign=m, keepaspectratio,]{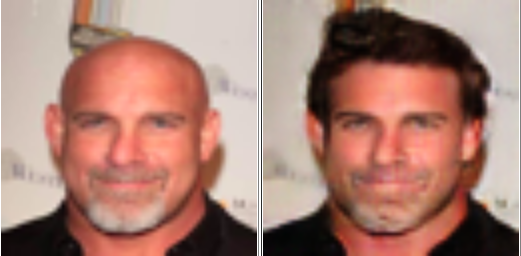}&
\includegraphics[width=0.29\textwidth,  ,valign=m, keepaspectratio,]{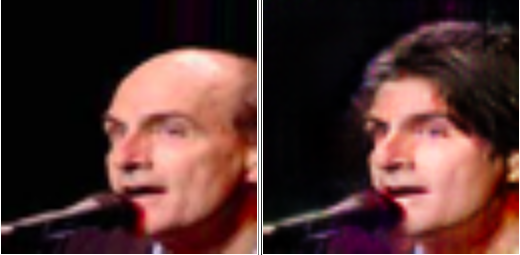}&
\includegraphics[width=0.29\textwidth,  ,valign=m, keepaspectratio,]{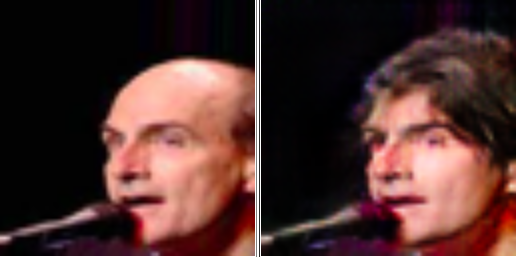}\\

\includegraphics[width=0.29\textwidth,  ,valign=m, keepaspectratio,]{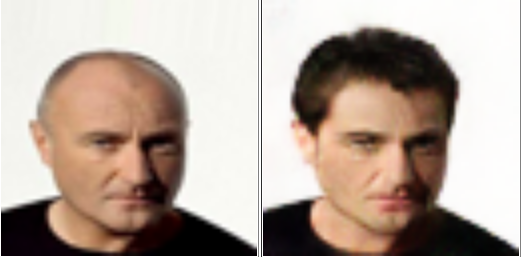} & 
\includegraphics[width=0.29\textwidth,  ,valign=m, keepaspectratio,]{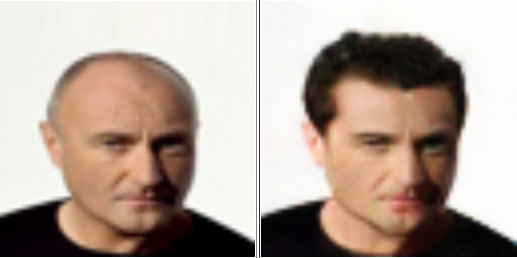}&
\includegraphics[width=0.29\textwidth,  ,valign=m, keepaspectratio,]{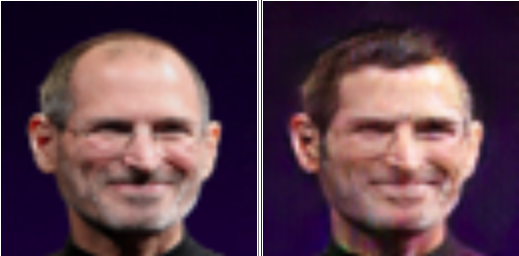}&
\includegraphics[width=0.29\textwidth,  ,valign=m, keepaspectratio,]{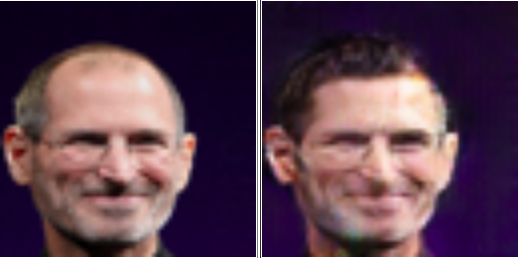}\\

\end{tabular}
\caption{Example results of \nth{3} experiment}
\end{figure*}

\begin{figure*}
\captionsetup[subfigure]{labelformat=empty}
\centering
\setlength\tabcolsep{0pt} 
\hspace*{-1.0cm}%
\begin{tabular}{cccc}
\hspace*{0.5cm}Black-Straight & \hspace*{0.4cm}Black-Wavy & \hspace*{0.4cm}Blond-Straight & \hspace*{0.3cm}Blond-Wavy\\

\includegraphics[width=0.29\textwidth,  ,valign=m, keepaspectratio,]{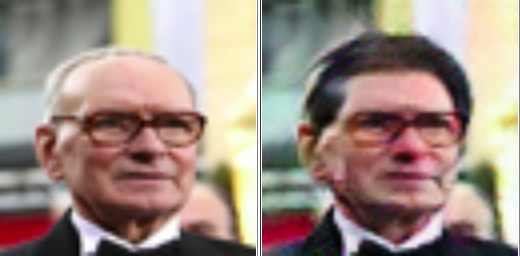} & 
\includegraphics[width=0.29\textwidth,  ,valign=m, keepaspectratio,]{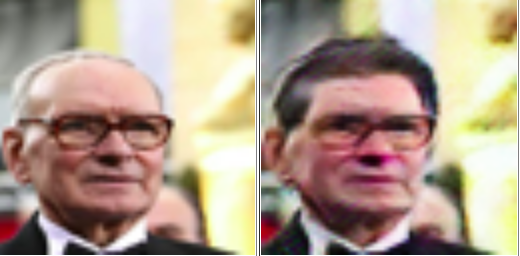}&
\includegraphics[width=0.29\textwidth,  ,valign=m, keepaspectratio,]{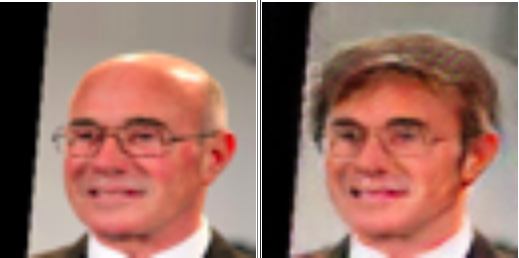}&
\includegraphics[width=0.29\textwidth,  ,valign=m, keepaspectratio,]{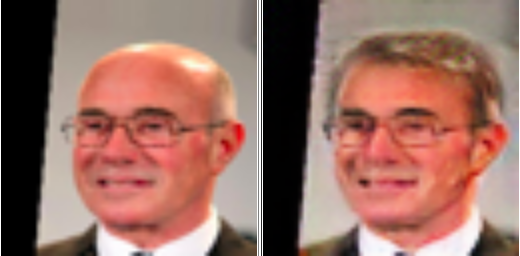}\\

\includegraphics[width=0.29\textwidth,  ,valign=m, keepaspectratio,]{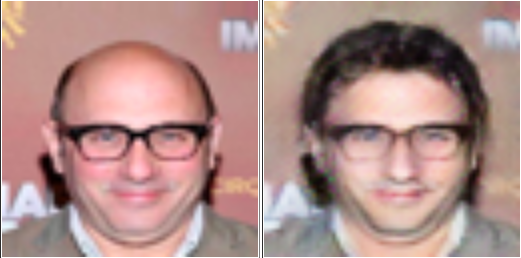} & 
\includegraphics[width=0.29\textwidth,  ,valign=m, keepaspectratio,]{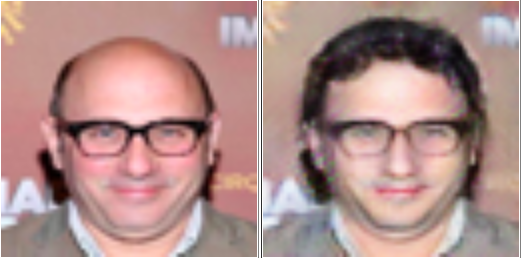}&
\includegraphics[width=0.29\textwidth,  ,valign=m, keepaspectratio,]{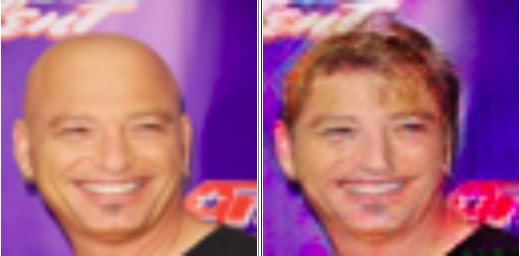}&
\includegraphics[width=0.29\textwidth,  ,valign=m, keepaspectratio,]{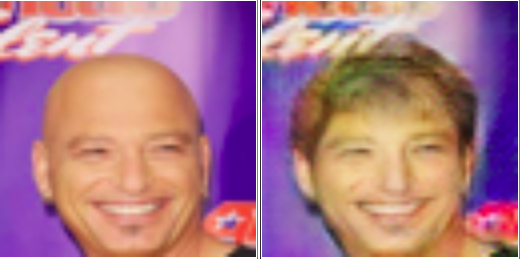}\\

\end{tabular}
\caption{Example results of \nth{4} experiment}
\end{figure*}

\begin{figure*}
\captionsetup[subfigure]{labelformat=empty}
\centering
\setlength\tabcolsep{0pt} 
\hspace*{-1.0cm}%
\begin{tabular}{cccc}
\hspace*{0.5cm}Black-Straight & \hspace*{0.4cm}Black-Wavy & \hspace*{0.4cm}Blond-Straight & \hspace*{0.3cm}Blond-Wavy\\

\includegraphics[width=0.29\textwidth,  ,valign=m, keepaspectratio,]{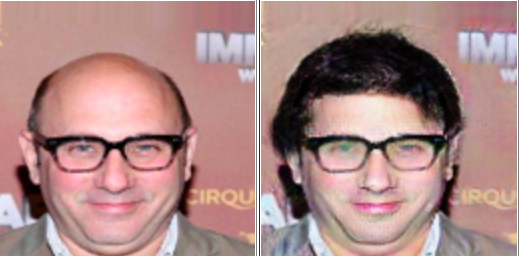} & 
\includegraphics[width=0.29\textwidth,  ,valign=m, keepaspectratio,]{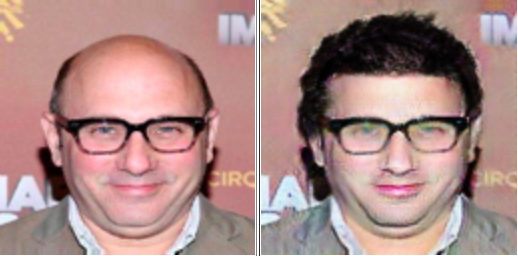}&
\includegraphics[width=0.29\textwidth,  ,valign=m, keepaspectratio,]{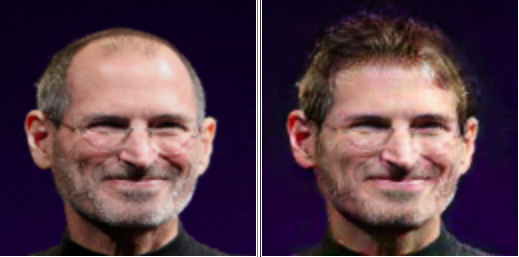}&
\includegraphics[width=0.29\textwidth,  ,valign=m, keepaspectratio,]{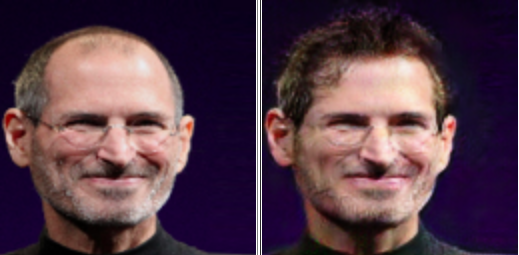}\\

\includegraphics[width=0.29\textwidth,  ,valign=m, keepaspectratio,]{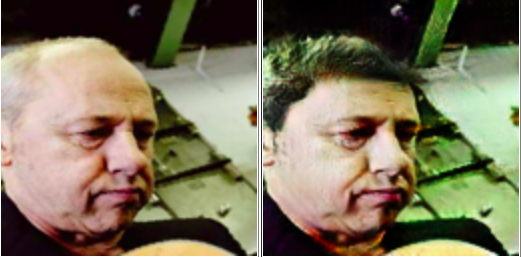} & 
\includegraphics[width=0.29\textwidth,  ,valign=m, keepaspectratio,]{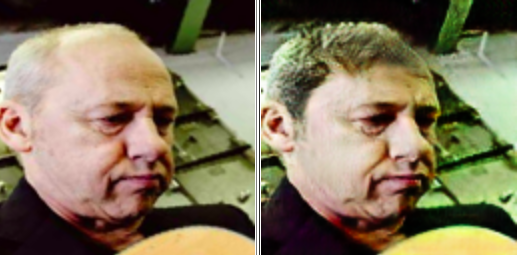}&
\includegraphics[width=0.29\textwidth,  ,valign=m, keepaspectratio,]{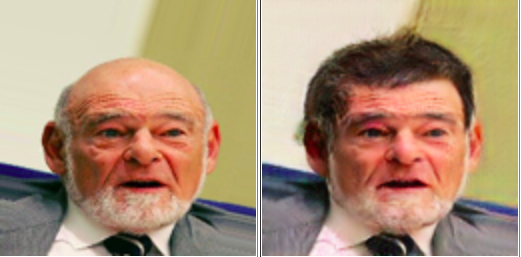}&
\includegraphics[width=0.29\textwidth,  ,valign=m, keepaspectratio,]{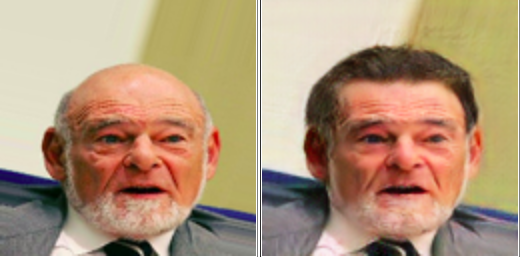}\\

\end{tabular}
\caption{Example results of \nth{5} experiment}
\end{figure*}

\begin{figure*}
\captionsetup[subfigure]{labelformat=empty}
\centering
\setlength\tabcolsep{0pt} 
\hspace*{-1.0cm}%
\begin{tabular}{cccc}
\hspace*{0.5cm}Black-Straight & \hspace*{0.4cm}Black-Wavy & \hspace*{0.4cm}Blond-Straight & \hspace*{0.3cm}Blond-Wavy\\

\includegraphics[width=0.29\textwidth,  ,valign=m, keepaspectratio,]{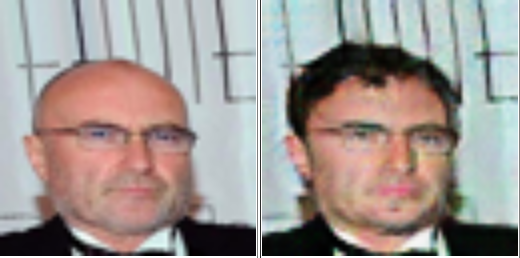} & 
\includegraphics[width=0.29\textwidth,  ,valign=m, keepaspectratio,]{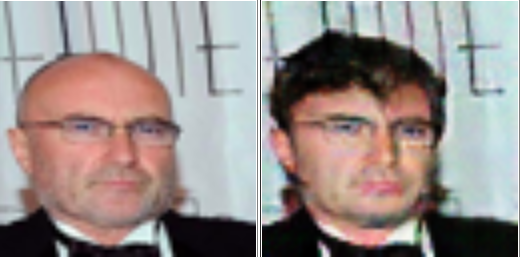}&
\includegraphics[width=0.29\textwidth,  ,valign=m, keepaspectratio,]{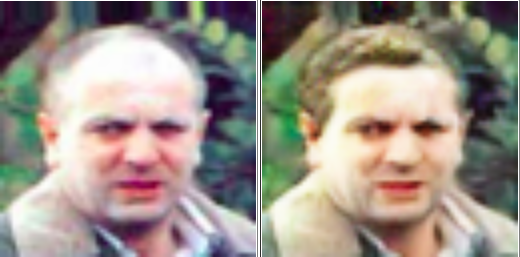}&
\includegraphics[width=0.29\textwidth,  ,valign=m, keepaspectratio,]{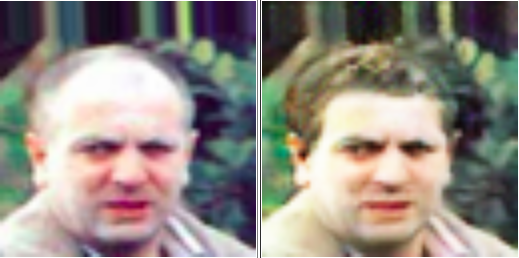}\\

\hspace*{0.5cm}Brown-Straight & \hspace*{0.4cm}Brown-Wavy & \hspace*{0.4cm}Gray-Straight & \hspace*{0.3cm}Gray-Wavy\\

\includegraphics[width=0.29\textwidth,  ,valign=m, keepaspectratio,]{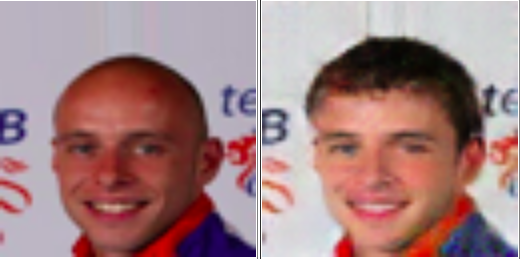} & 
\includegraphics[width=0.29\textwidth,  ,valign=m, keepaspectratio,]{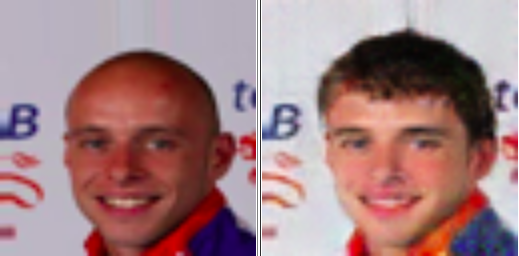}&
\includegraphics[width=0.29\textwidth,  ,valign=m, keepaspectratio,]{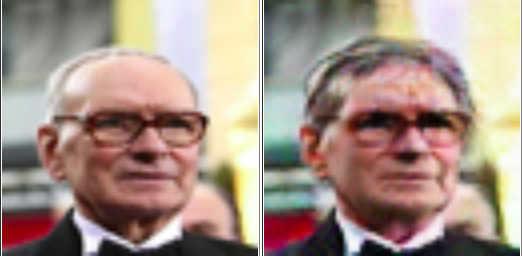}&
\includegraphics[width=0.29\textwidth,  ,valign=m, keepaspectratio,]{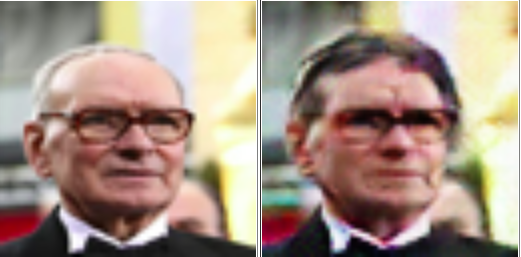}\\

\end{tabular}
\caption{Example results of \nth{6} experiment}
\end{figure*}

\begin{figure*}
\captionsetup[subfigure]{labelformat=empty}
\centering
\setlength\tabcolsep{0pt} 
\hspace*{-1.0cm}%
\begin{tabular}{cccc}
\hspace*{0.5cm}Black-Straight & \hspace*{0.4cm}Black-Wavy & \hspace*{0.4cm}Blond-Straight & \hspace*{0.3cm}Blond-Wavy\\

\includegraphics[width=0.29\textwidth,  ,valign=m, keepaspectratio,]{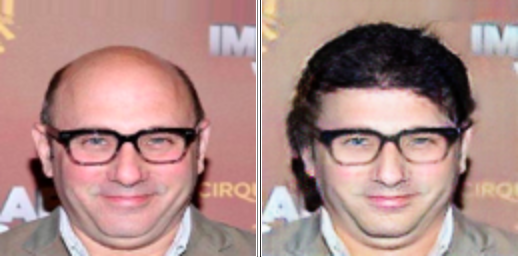} & 
\includegraphics[width=0.29\textwidth,  ,valign=m, keepaspectratio,]{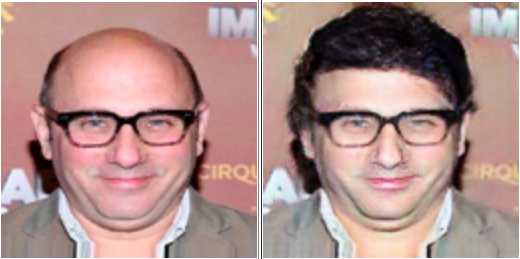}&
\includegraphics[width=0.29\textwidth,  ,valign=m, keepaspectratio,]{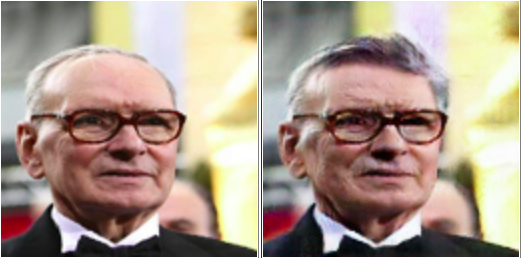}&
\includegraphics[width=0.29\textwidth,  ,valign=m, keepaspectratio,]{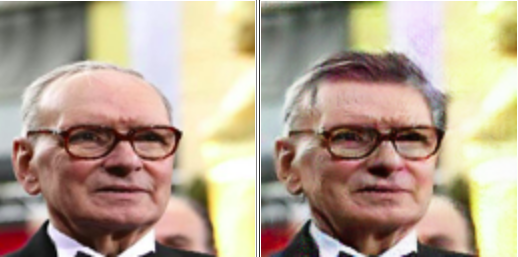}\\

\hspace*{0.5cm}Brown-Straight & \hspace*{0.4cm}Brown-Wavy & \hspace*{0.4cm}Gray-Straight & \hspace*{0.3cm}Gray-Wavy\\

\includegraphics[width=0.29\textwidth,  ,valign=m, keepaspectratio,]{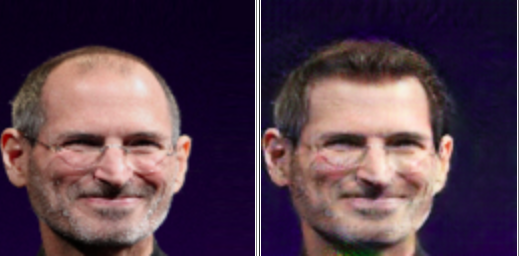} & 
\includegraphics[width=0.29\textwidth,  ,valign=m, keepaspectratio,]{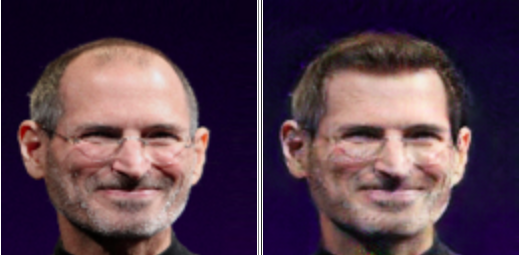}&
\includegraphics[width=0.29\textwidth,  ,valign=m, keepaspectratio,]{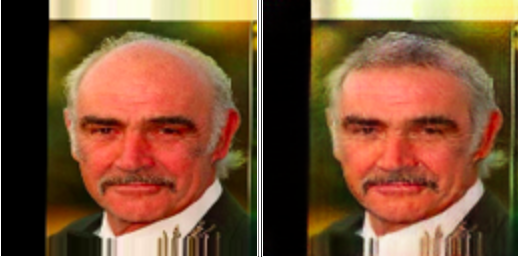}&
\includegraphics[width=0.29\textwidth,  ,valign=m, keepaspectratio,]{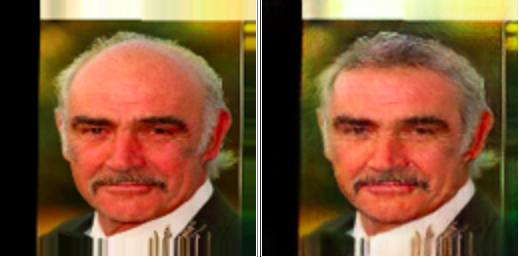}\\

\end{tabular}
\caption{Example results of \nth{7} experiment}
\end{figure*}

\begin{figure*}
\captionsetup[subfigure]{labelformat=empty}
\centering
\setlength\tabcolsep{0pt} 
\hspace*{-1.0cm}%
\begin{tabular}{cccc}
\hspace*{0.5cm}Black-Straight & \hspace*{0.4cm}Black-Wavy & \hspace*{0.4cm}Blond-Straight & \hspace*{0.3cm}Blond-Wavy\\

\includegraphics[width=0.29\textwidth,  ,valign=m, keepaspectratio,]{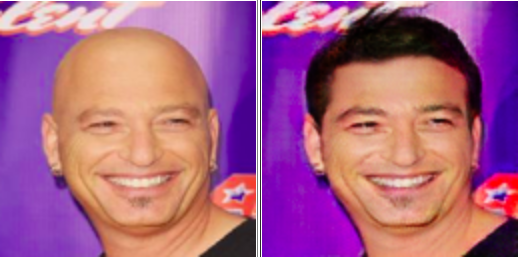} & 
\includegraphics[width=0.29\textwidth,  ,valign=m, keepaspectratio,]{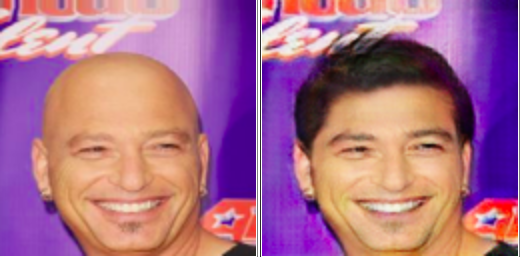}&
\includegraphics[width=0.29\textwidth,  ,valign=m, keepaspectratio,]{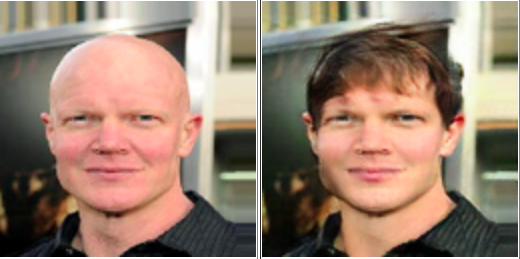}&
\includegraphics[width=0.29\textwidth,  ,valign=m, keepaspectratio,]{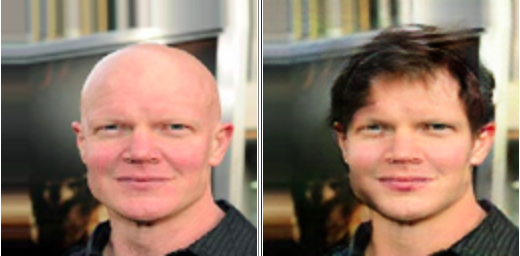}\\

\includegraphics[width=0.29\textwidth,  ,valign=m, keepaspectratio,]{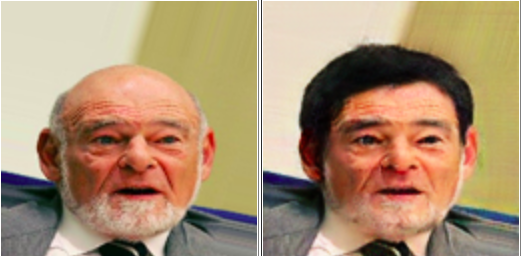} & 
\includegraphics[width=0.29\textwidth,  ,valign=m, keepaspectratio,]{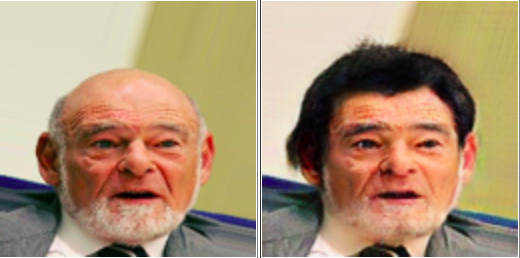}&
\includegraphics[width=0.29\textwidth,  ,valign=m, keepaspectratio,]{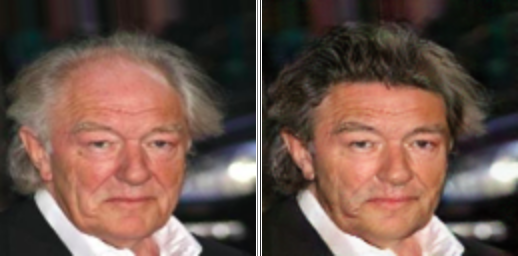}&
\includegraphics[width=0.29\textwidth,  ,valign=m, keepaspectratio,]{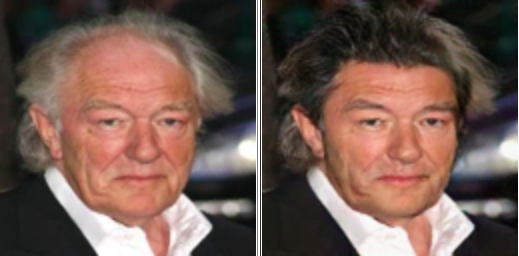}\\

\end{tabular}
\caption{Example results of \nth{8} experiment}
\end{figure*}

\begin{figure*}
\captionsetup[subfigure]{labelformat=empty}
\centering
\setlength\tabcolsep{0pt} 
\hspace*{-1.0cm}%
\begin{tabular}{cccc}
\hspace*{0.5cm}Black-Straight & \hspace*{0.4cm}Black-Wavy & \hspace*{0.4cm}Blond-Straight & \hspace*{0.3cm}Blond-Wavy\\

\includegraphics[width=0.29\textwidth,  ,valign=m, keepaspectratio,]{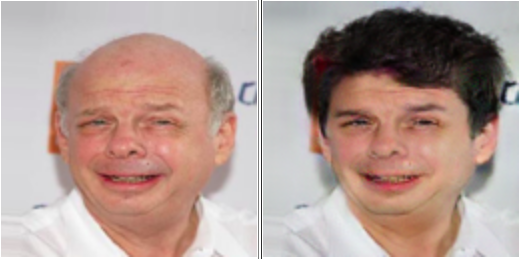} & 
\includegraphics[width=0.29\textwidth,  ,valign=m, keepaspectratio,]{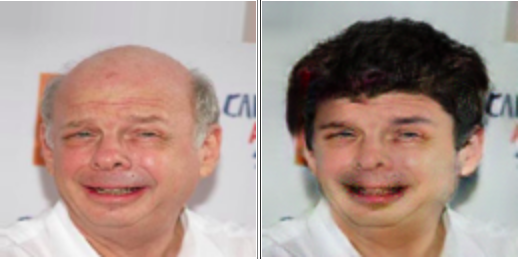}&
\includegraphics[width=0.29\textwidth,  ,valign=m, keepaspectratio,]{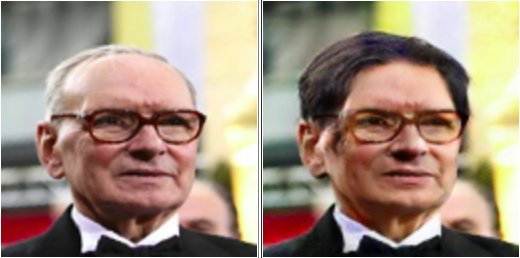}&
\includegraphics[width=0.29\textwidth,  ,valign=m, keepaspectratio,]{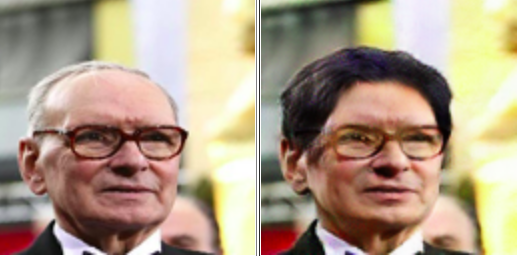}\\

\includegraphics[width=0.29\textwidth,  ,valign=m, keepaspectratio,]{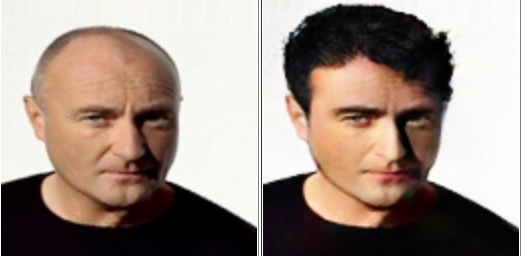} & 
\includegraphics[width=0.29\textwidth,  ,valign=m, keepaspectratio,]{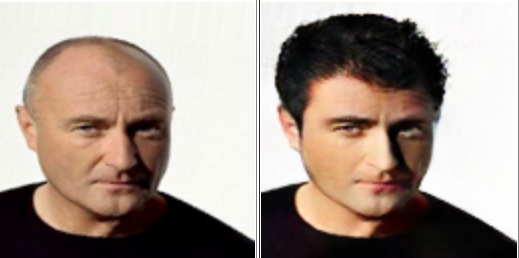}&
\includegraphics[width=0.29\textwidth,  ,valign=m, keepaspectratio,]{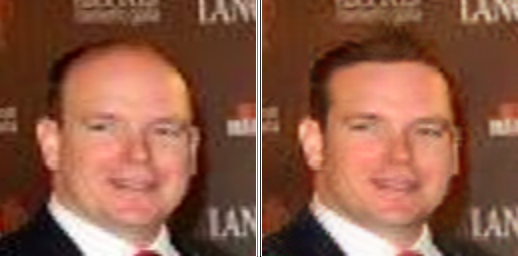}&
\includegraphics[width=0.29\textwidth,  ,valign=m, keepaspectratio,]{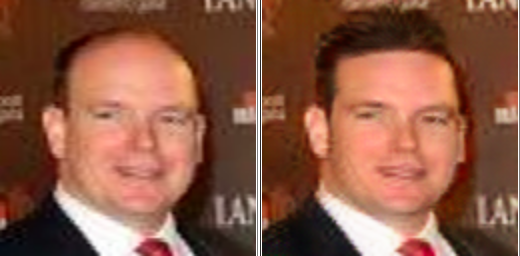}\\

\end{tabular}
\caption{Example results of \nth{9} experiment}
\end{figure*}

\begin{figure*}
\captionsetup[subfigure]{labelformat=empty}
\centering
\setlength\tabcolsep{0pt} 
\hspace*{-1.0cm}%
\begin{tabular}{cccc}

\includegraphics[width=0.29\textwidth,  ,valign=m, keepaspectratio,]{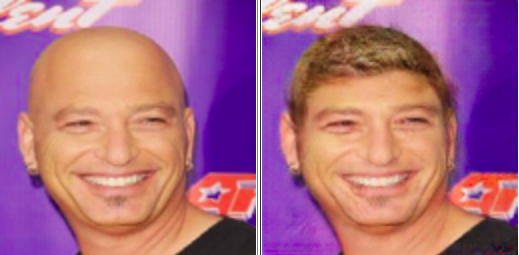} & 
\includegraphics[width=0.29\textwidth,  ,valign=m, keepaspectratio,]{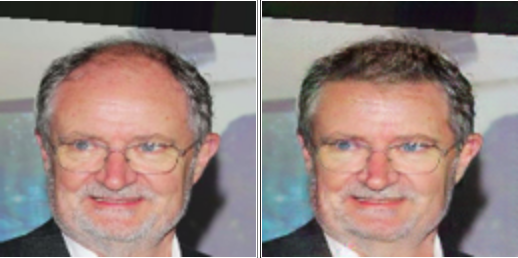}&
\includegraphics[width=0.29\textwidth,  ,valign=m, keepaspectratio,]{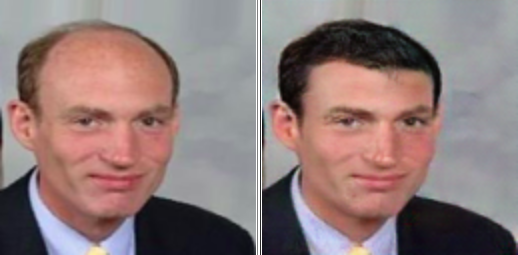}&
\includegraphics[width=0.29\textwidth,  ,valign=m, keepaspectratio,]{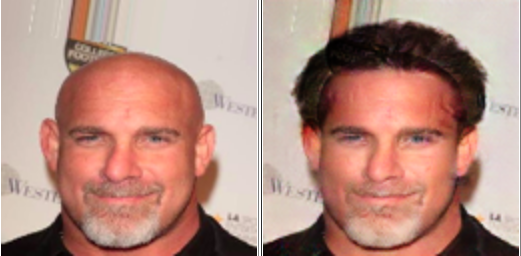}\\

\includegraphics[width=0.29\textwidth,  ,valign=m, keepaspectratio,]{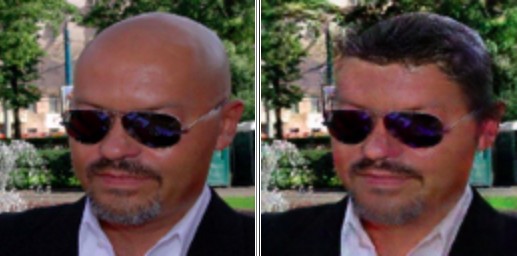} & 
\includegraphics[width=0.29\textwidth,  ,valign=m, keepaspectratio,]{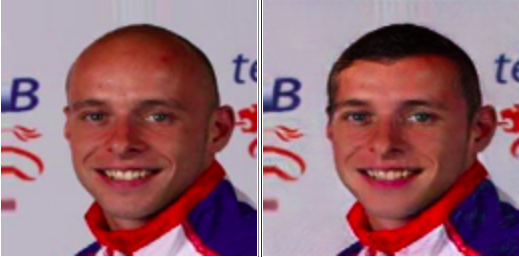}&
\includegraphics[width=0.29\textwidth,  ,valign=m, keepaspectratio,]{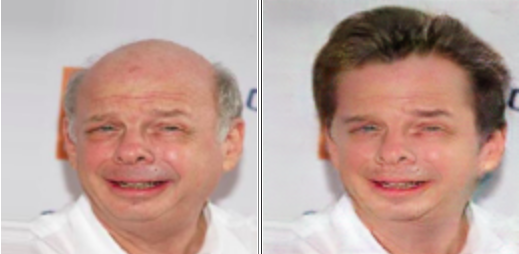}&
\includegraphics[width=0.29\textwidth,  ,valign=m, keepaspectratio,]{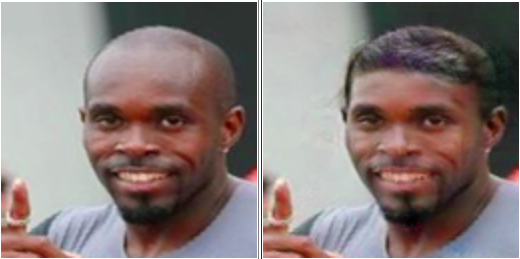}\\

\end{tabular}
\caption{Example results of \nth{10} experiment (Note that there is no condition layer in this experiment)}
\end{figure*}

\begin{figure*}
\captionsetup[subfigure]{labelformat=empty}
\centering
\setlength\tabcolsep{0pt} 
\hspace*{-1.0cm}%
\begin{tabular}{cccc}
\hspace*{0.5cm}Black-Straight & \hspace*{0.4cm}Black-Wavy & \hspace*{0.4cm}Blond-Straight & \hspace*{0.3cm}Blond-Wavy\\

\includegraphics[width=0.29\textwidth,  ,valign=m, keepaspectratio,]{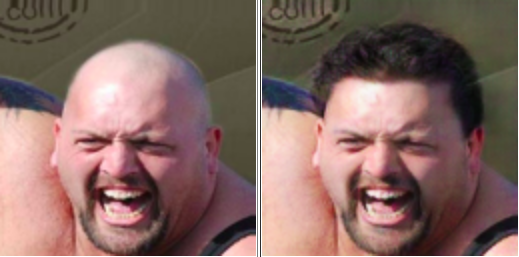} & 
\includegraphics[width=0.29\textwidth,  ,valign=m, keepaspectratio,]{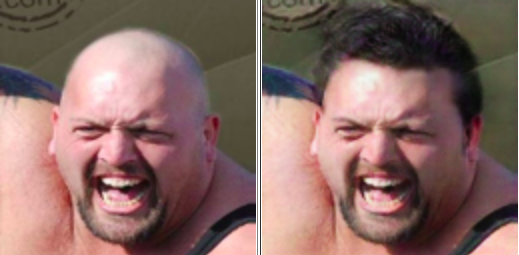}&
\includegraphics[width=0.29\textwidth,  ,valign=m, keepaspectratio,]{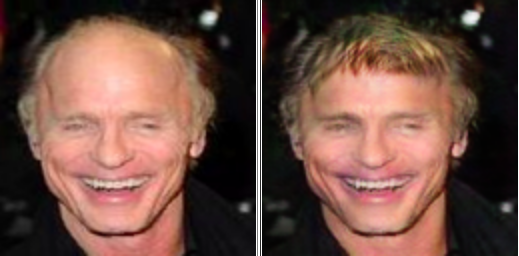}&
\includegraphics[width=0.29\textwidth,  ,valign=m, keepaspectratio,]{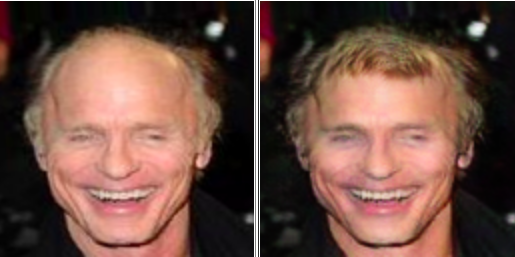}\\

\includegraphics[width=0.29\textwidth,  ,valign=m, keepaspectratio,]{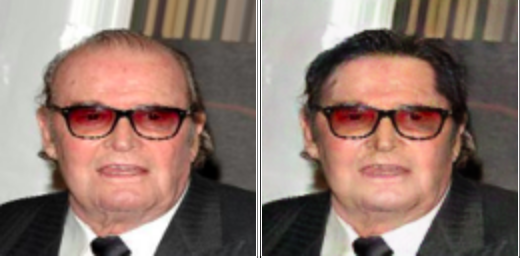} & 
\includegraphics[width=0.29\textwidth,  ,valign=m, keepaspectratio,]{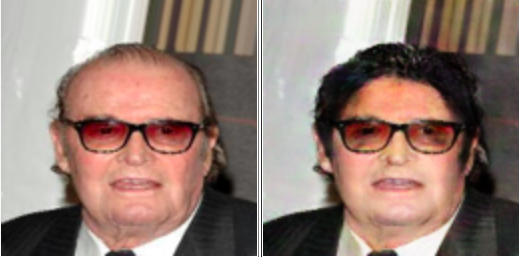}&
\includegraphics[width=0.29\textwidth,  ,valign=m, keepaspectratio,]{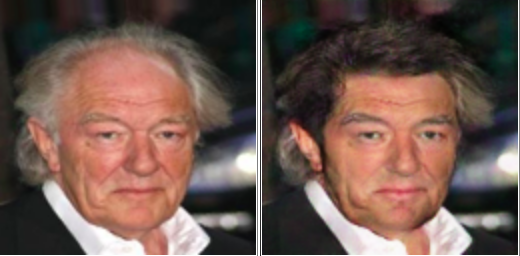}&
\includegraphics[width=0.29\textwidth,  ,valign=m, keepaspectratio,]{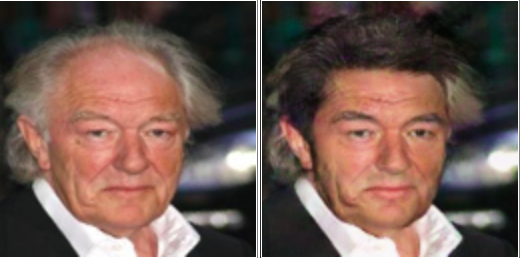}\\

\end{tabular}
\caption{Example results of \nth{11} experiment}
\end{figure*}

\begin{figure*}
\captionsetup[subfigure]{labelformat=empty}
\centering
\setlength\tabcolsep{0pt} 
\hspace*{-1.0cm}%
\begin{tabular}{cccc}
\hspace*{0.5cm}Black-Straight & \hspace*{0.4cm}Black-Wavy & \hspace*{0.4cm}Blond-Straight & \hspace*{0.3cm}Blond-Wavy\\

\includegraphics[width=0.29\textwidth,  ,valign=m, keepaspectratio,]{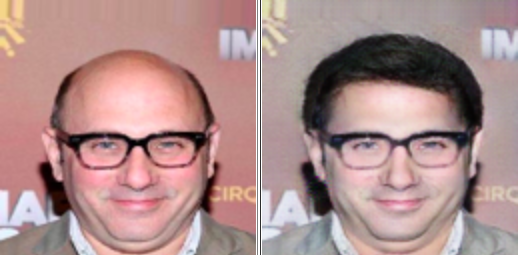} & 
\includegraphics[width=0.29\textwidth,  ,valign=m, keepaspectratio,]{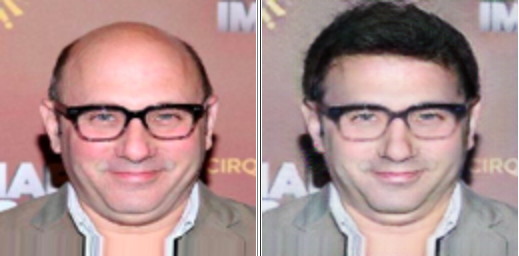}&
\includegraphics[width=0.29\textwidth,  ,valign=m, keepaspectratio,]{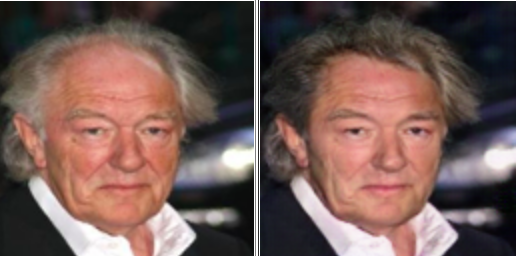}&
\includegraphics[width=0.29\textwidth,  ,valign=m, keepaspectratio,]{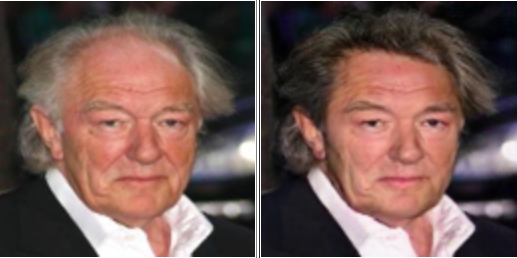}\\

\includegraphics[width=0.29\textwidth,  ,valign=m, keepaspectratio,]{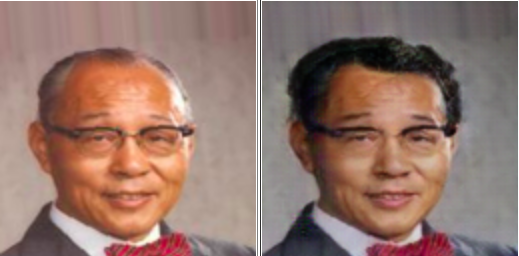} & 
\includegraphics[width=0.29\textwidth,  ,valign=m, keepaspectratio,]{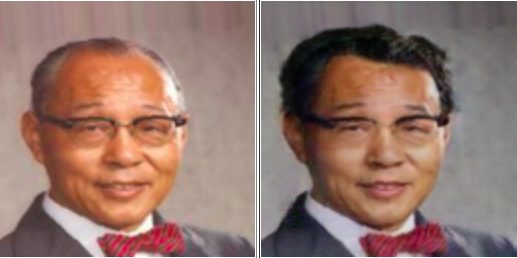}&
\includegraphics[width=0.29\textwidth,  ,valign=m, keepaspectratio,]{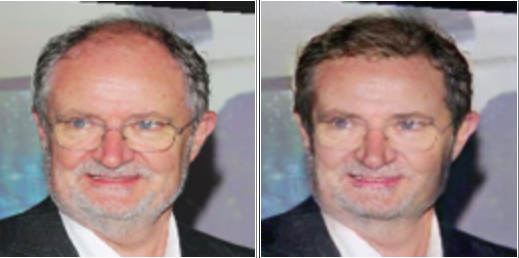}&
\includegraphics[width=0.29\textwidth,  ,valign=m, keepaspectratio,]{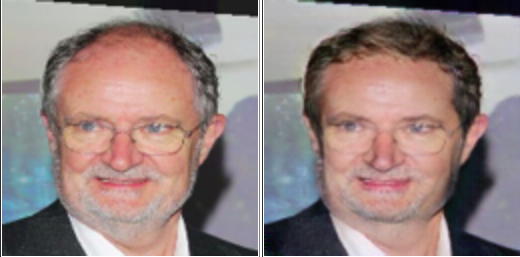}\\

\end{tabular}
\caption{Example results of \nth{12} experiment}
\end{figure*}

{\small
\bibliographystyle{ieee_fullname}
\bibliography{egbib}
}

\end{document}